\documentclass{interact}

\usepackage[utf8]{inputenc}
\usepackage{graphicx}
\usepackage{url}
\graphicspath{ {images/} }
\usepackage{lmodern}
\usepackage{amsmath}
\usepackage{verbatim}

\usepackage{listings}
\usepackage{subfig}
\usepackage{fancyvrb}

\usepackage{threeparttable} 

\DeclareUnicodeCharacter{03BC}{\ensuremath{\mu}}

\usepackage{hyperref}
\hypersetup{
    colorlinks=true,
    linkcolor=blue,
    citecolor=blue,
    filecolor=magenta,      
    urlcolor=cyan,
    pdftitle={Overleaf Example},
    pdfpagemode=FullScreen
    }
  
\usepackage{orcidlink}

\usepackage{natbib}
\bibpunct[, ]{(}{)}{;}{a}{}{,}
\renewcommand\bibfont{\fontsize{10}{12}\selectfont}
\begin{document}

\title{Enhancing kelp forest detection in remote sensing images using crowdsourced labels with Mixed Vision Transformers and ConvNeXt segmentation models}







\author{
\name{Ioannis Nasios\textsuperscript{a}\thanks{CONTACT Ioannis Nasios.
Email: ioannis.nasios@nodalpoint.com} \orcidlink{0000-0002-1765-0646}}
\affil{\textsuperscript{a}Nodalpoint Systems, Pireos 205, 118 53, Athens, Greece} }

\maketitle

\begin{abstract}

Kelp forests, as foundation species, are vital to marine ecosystems, providing essential food and habitat for numerous organisms. This study explores the integration of crowdsourced labels with advanced artificial intelligence models to develop a fast and accurate kelp canopy detection pipeline using Landsat images. Building on the success of a machine learning competition, where this approach ranked third and performed consistently well on both local validation and public and private leaderboards, the research highlights the effectiveness of combining Mixed Vision Transformers (MIT) with ConvNeXt models. Training these models on various image sizes significantly enhanced the accuracy of the ensemble results. U-Net emerged as the best segmentation architecture, with UpperNet also contributing to the final ensemble. Key Landsat bands, such as ShortWave InfraRed (SWIR1) and Near-InfraRed (NIR), were crucial while altitude data was used in postprocessing to eliminate false positives on land. The methodology achieved a high detection rate, accurately identifying about three out of four pixels containing kelp canopy while keeping false positives low. Despite the medium resolution of Landsat satellites, their extensive historical coverage makes them effective for studying kelp forests. This work also underscores the potential of combining machine learning models with crowdsourced data for effective and scalable environmental monitoring. All running code for training all models and inference can be found at \url{https://github.com/IoannisNasios/Kelp_Forests}.

\end{abstract}

\begin{keywords}   
kelp forests; segmentation; artificial intelligence; crowdsourcing; infrared; remote sensing;  
\end{keywords}




\section{Introduction}
\label{sec:introduction}



Kelp forests are highly productive components of cold water that are found in rocky marine coastlines globally (roughly in the 25\% of the world's coastlines). These are physiologically constrained by light at high latitudes and by nutrients and warm temperatures at low latitudes. Within mid latitude belts, well developed kelp forests are most threatened by herbivores, usually from sea urchins. Overfishing and extirpation of highly valued predators often triggered herbivore population increases, leading to widespread kelp deforestation \citep{steneck2002kelp}. As it was not clear whether it was nutrient availability or grazers (sea urchins) that had the most influence over kelp forest health, size and longevity, after using Landsat imagery to look at long-term trends, and comparing those trends to known differences between Central and Southern California waters, \citet{cavanaugh2011environmental} found that it was a third force affecting kelp dynamics, the wave disturbance. Strong waves generated by storms uproot the kelp from their holdfasts and can devastate the forests far more than any grazer. Furthermore, climate-driven increases in storm frequency simplify kelp forest food webs as species go locally extinct \citep{byrnes2011climate} while climate change impacts on kelp forests for a variety of reasons has risen sharply \citep{smale2020impacts}. Kelp is a foundation species, as it provides food for diverse types of herbivores from tiny shrimps to ravenous sea urchins to grazing fish. Giant kelp particularly is extraordinary as it has one of the widest global distributions and is one of the easiest to see from space even using medium resolution satellites as Landsat. Over the last years, many have used satellite data to study kelp forests distribution and characteristics. To tease apart signals due to anthropogenic effect from natural variability, a long-term analysis trends is critical. \citet{bell2020three} used three decades of Landsat data to study the variability in California's giant kelp forests, while \citet{filbee2019arctic} studied their diversity, resilience and future.

Giant kelp forests are among Earth's most productive habitats, and their great diversity of plant and animal species supports many fisheries around the world. They are the world's largest marine plants and regularly grow up to 35 meters tall. They live for seven years at most, and often they disappear before that because of winter storms or over-grazing by other species. 
Given the right balance of conditions, giant kelp can grow as much as 50 centimeters per day, and this robust growth makes it possible for kelp fronds to be commercially harvested. 
Today, only a few thousand tons of giant kelp are harvested each year, some by hand and some by mechanical harvesters. The kelp can be trimmed no lower than 4 feet below the water surface for harvesting to be sustainable. Studies have shown that negative affects are negligible, although some fish populations are temporarily displaced, \citep{FloatingForests}.

As kelp forests comprise one of the most important earth's ecosystems, mapping and monitoring them is essential. Several methods have been developed over the last years as earth observation tools have been improved and many are freely available. \citet{mora2020high} have proposed a method for high resolution kelp mapping with sentinel-2 imagery using computational filters-indexes. Also \citet{gendall2023multi} proposed a multi-satellite mapping framework that leverage the monitoring of floating kelp forest ecosystems using medium-resolution imagery from the 1970s onwards, and more recently, using high-resolution imagery from the early 2000s onwards. Satellite imagery enables the mapping of existing and historical giant kelp populations in understudied regions, but automating the detection of giant kelp using satellite imagery requires approaches that are robust to the optical complexity of the shallow, nearshore environment. \citet{houskeeper2022automated} studied the automated satellite remote sensing of giant kelp at the Falkland islands which is also the study area of current research.

Various segmentation models, provided in python libraries, are built upon the pytorch framework. Among the most popular is the segmentation\_models\_pytorch (SMP) \citep{Iakubovskii:2019} which is also very easy to use. SMP provides easy access to many encoder models and also easy access to the timm encoder models \citep{rw2019timm}. Furthermore, SMP has implemented some famous segmentation architectures which can be combined with the provided encoders. Transformer models first developed for Natural Language Processing (NLP) tasks but over the last years are widely used in computer vision tasks as well. The UpperNet segmentation architecture which is also used in our here, is included in the transformers library \citep{wolf-etal-2020-transformers} along with various ConvNeXt or Swin encoders (130k stars on github, many stars due to NLP). 
Both SMP and transformers libraries were used for the current research successfully in a straight forward approach easy to understand and reproduce. \autoref{table:pop} shows the popularity of these common used for segmentation python libraries as this is reflected by number of stars on github (August 08 2024).

\begin{table}[ht!]
\renewcommand{\arraystretch}{1.3}
\caption{Pytorch libraries popularity}
\label{table:pop}
\centering
\begin{tabular}{l  c  } 
 \hline
 Library & github stars  \\
 \hline
segmentation\_models\_pytorch  & 9300  \\
timm                           & 31000  \\
transformers                   & 130000  \\
\hline

\end{tabular}
\end{table}

Nowadays, artificial intelligence methods has dominated many scientific areas with computer vision being one of the most explored and exploited domains. Semantic segmentation is the computer vision task which is used to categorize each pixel of an image into a class. It has a broad range of applications in a variety of domains including earth observation, autonomous driving and medical image analysis. Semantic segmentation in earth observation is a demanding and fast growing area as new machine learning models and new satellites are becoming available frequently. This can be used as a cost-effective and scalable solution to understand and protect kelp forest dynamics. \citet{he2022swin} proposed the Swin transformer encoder over a U-Net segmentation architecture as a state-of-the-art approach to earth observation data but as from current research combining transformers with convolution outperforms the use of transformers alone. 

Proposed models here have been used separated before, on various segmentation or other encoder-decoder tasks. \citet{thisanke2023semantic} at their survey, compared and gave prominence of the power of the transformer based models in segmentation tasks, while \citet{roy2023mednext} used transformer-driven scaling on ConvNets for semantic segmentation of medical images and \citet{ji2024rethinking} used vision transformers for polyp segmentation with an out-of-distribution perspective. \citet{zhang2023practical} combined in a single model the power of swin transformes and convolution models for image denoising using a Swin-Conv-UNet and data synthesis and also \citet{li2023depthformer} in Depthformer combined two encoders in one model, a transformer and a convolutional one, for accurate monocular depth estimation.
After testing various state-of-the-art models and model combinations, the combination of Transformer and ConvNeXt models proved to be the most effective, offering greater accuracy and suitability for semantic segmentation in earth observation data.


Machine learning competitions are significant contributors in advancing the machine learning field that can add a different perspective parallel to the traditional academic. In a competition, state-of-the-art models and training strategies are evaluated and refined, demanding creativity, innovation, and other key qualities. Many machine learning competition platforms have been developed with kaggle, drivendata and AIcrowd being some of the most important \citep{carlens2024state}. Competitions held within small, closed circles, rather than on large platforms, often yield less impressive results due to their limited competitiveness. Sharing insights gained from a major machine learning competition with the research community fosters the advancement of new knowledge and ideas, both related to the specific competition topic and in a broader context \citep{nasios2024analyze, nasios2022blending}. This research leverages a recent competition to present a high-performing method applied to an original segmentation dataset. It also discusses the insights gained during the process, along with perspectives, potential applications, and suggestions for future improvement. 

Additionally, the process provided valuable insights, along with perspectives, potential use cases, and suggestions for further improvement, all of which are discussed.

The main satellite data comes from the level 2 Landsat 5, 7 and 8 imagery.
\autoref{table:landsat_run} shows the period that each satellite was active and produced satellite products. Using Level-2 data can be very important as these are ready to be used without any corrections or intercalibrating between different satellite data sets need to be applied. All 3 together have a time span of 40 years which makes them ideal for long period studies and for monitoring of structures, detectable under their 30m resolution, as are the giant kelp forests. The ability of satellites to capture not only light in the visible wavelengths but also in the infrared bands, the bands that kelp forests are mostly detectable, enhances the success of this task. Except from the Landsat satellite data an extra band was provided, containing the Digital Elevation Model (DEM) which is generated from the the Terra Advanced Spaceborne Thermal Emission and Reflection Radiometer (ASTER) that can be used to generate a land-sea mask in approximately the same resolution. 

\begin{table}[ht!]
\renewcommand{\arraystretch}{1.3}
\caption{Landsat satellites periods}
\label{table:landsat_run}
\centering
\begin{tabular}{l  c  c } 
 \hline
 Landsat & Run from & Run until  \\
 \hline
Landsat 5  & 1984  & 2013  \\
Landsat 7  & 1999  & 2022  \\
Landsat 8  & 2013  & today \\
\hline
\end{tabular}
\end{table}

Landsat satellites have been used in various research over the years. \citet{el2014change} studied the change detection of coral reef habitat using Landsat 5,  7 and Landsat 8 OLI data in the Red Sea. \citet{zaghian2023enhancing} found that suspended sediment concentration retrieval can be enhanced upon integrating thermal infrared and optical bands of Landsat-8 with machine learning algorithms. \citet{finger2021mapping} studied the mapping of bull kelp canopy in northern California using Landsat to enable long-term monitoring while \citet{simms2001satellite} through the comparative analysis of HRV and Thematic Mapper, estimated the seasonal variation of the submerged kelp beds biomass on the Atlantic coast of Canada.

Citizen scientists play diverse and significant roles in earth observation. \citet{boyd2022citizen} explored the use of citizen science for earth observation within the UK, while \citet{karagiannopoulou2022data} reviewed various applications, methods, and future trends in data fusion for earth observation, focusing on the role of citizens as sensors. One of the most important contributions of citizen science is crowdsourced data labeling, which serves as an alternative to expert annotations. This approach has been widely adopted to create essential datasets for training AI systems. By involving many participants in the annotation process, large datasets can be effectively labeled and categorized, with label uncertainty minimized through multiple evaluations of the same data. Furthermore,  combining crowdsourcing with deep learning led to higher accuracy and reduced volunteer effort in mapping human settlements, \citet{herfort2019mapping}.

The study presents an algorithm developed to predict the presence or absence of kelp forests using satellite imagery. Given the relatively small dataset and the limited geographical distribution of kelp forests, detection is challenging. Furthermore, the 30-meter resolution of Landsat imagery poses extra challenges for detecting the smaller kelp canopies. Additionally, since citizen scientists contributed to the labeling, some noise was introduced into the dataset, highlighting the need for a robust, standardized approach. Developing a reliable high performing methodology for kelp forest detection is essential for ongoing monitoring of these unique ecosystems, enabling researchers to study their dynamics and the factors influencing their distribution over time. Such a tool could provide critical insights for predicting future changes in kelp forests and for supporting proactive conservation efforts.

\section{Methodology}
\label{sec:Methodology}
Current research is based on a machine learning competition in which the author participated and exploited. The competition \href{https://www.drivendata.org/competitions/255/kelp-forest-segmentation/}{Kelp Wanted: Segmenting Kelp Forests} was a semantic segmentation competition with Landsat satellite data, aiming in mapping kelp forests areas. The competition was sponsored by \href{https://www.mathworks.com/}{Mathworks} and included partners as \href{https://byrneslab.net/}{the Byrnes Lab of UMass Boston} and the \href{https://www.whoi.edu}{Woods Hole Oceanographic Institution} which also provided the raw data. Desideratum was a machine learning model that can be used around the planet to swiftly and efficiently classify giant kelp with at least the same accuracy as the human eye.


 

\subsection{Data}
\label{sec:Data}
The provided dataset was consisted by the feature data, meaning the satellite imagery from the Landsat satellite missions, the elevation information and the cloud masks. There was also the label data, the binary masks indicating the presence or absence of kelp canopy which were created by citizen scientists via the \href{https://www.zooniverse.org/projects/zooniverse/floating-forests}{Floating Forests platform}. Details for the creation of these kelp labels, such as how many citizens should have marked the kelp canopy to be set as kelp, was not shared. For competition purposes, a test dataset containing only satellite images (without labels) was also provided. Scoring on this dataset was available exclusively through the competition leaderboard.

The training dataset is consisted of 5635 square images (chips) of size 350 (pixels) and corresponding kelp forests masks. This image size was selected as similar to the image size given to citizen scientist for labelling. As Landsat resolution is 30m, each chip covers an area of about 10.5*10.5 square Km. Each image contains 7 channels, 5 Landsat channels (Blue, Green, Red, NIR, SWIR1), 
1 channel containing a cloud mask indicating the presence or absence of clouds and 1 channel containing the Digital Elevation Model (DEM), with elevation measured in meters from sea-level, as derived from the Advanced Spaceborne Thermal Emission and Reflection Radiometer (ASTER) data, which is utilized to generate a land-sea mask. Images from Landsat 5, 7 and 8 was used to span multiple decades, for a comprehensive view of the region's kelp forests over time, but the information of the Landsat product that each chip came from and with this the timestamp it belongs to was not available. 




The most useful bands for vegetation in general and kelp forests in particulate are the SWIR1 band (Shortwave Infrared), which is useful for distinguishing between different types of vegetation as well as for detecting moisture content in soil and vegetation and the NIR band (Near-Infrared) as healthy vegetation reflects a significant amount of NIR light. Together with the Red, Green and Blue bands, capture a broad spectrum of electromagnetic radiation and have been selected by host for their usefulness in monitoring coastal environments. This also shown in various studies for detecting and analyzing vegetation characteristics as \citet{bartold2024estimating} also shown in plant stress detection at peatlands using Sentinel-2 satellite imagery. A plant leaf typically has a low reflectance in the visible spectral region because of strong absorption by chlorophyll, a relatively high reflectance in the near-infrared because of internal leaf scattering and no absorption, and a relatively low reflectance in the infrared beyond 1.3μ because of strong absorption by water \citep{knipling1970physical}.

\begin{figure}[h]
    \centering
    \captionsetup{width=.5\linewidth}
    \includegraphics[width=0.5\textwidth,trim=0 40 0 0, clip]{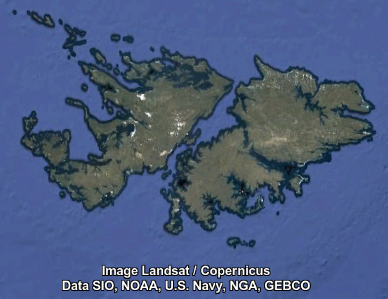}
    \caption{Falkland Islands }
    \label{fig:FalklandIslands}
\end{figure}

The dataset focuses on the coastal waters surrounding the Falkland Islands, covering multiple decades and providing a detailed view of changes in kelp forest distribution over time. The Falkland Islands (\autoref{fig:FalklandIslands}), an archipelago on the Patagonian shelf in the South Atlantic Ocean, are rich in kelp forests and ideal for this research, with an extensive coastline that supports abundant kelp growth. This region has been frequently imaged by Landsat over the years, resulting in numerous similar images, particularly of land and open sea areas without kelp, which reduces the diversity of the training data.

\subsection{Annotation errors}
\label{sec:Annotation errors}
Since the ground truth labels were provided by citizen scientists based on their visual interpretation of processed satellite images, there is some uncertainty regarding their accuracy. An ideal annotation should had been created by scuba divers or by dredging up samples from the deep. This would require a lot of effort to be collected and at the same time it would be limited to reachable places and would induce an increased cost. Fortunately, this important megastructures in the sea can be seen from space, alleviating the need for in situ sampling. There are although some false annotations which downgrades the quality of the produced models but as developed methodology here is robust, a curated dataset could significantly improve method's performance. 

\begin{figure}[h]
    \centering
    \captionsetup{width=.65\linewidth}
    \includegraphics[width=0.65\textwidth,trim=0 0 0 0, clip]{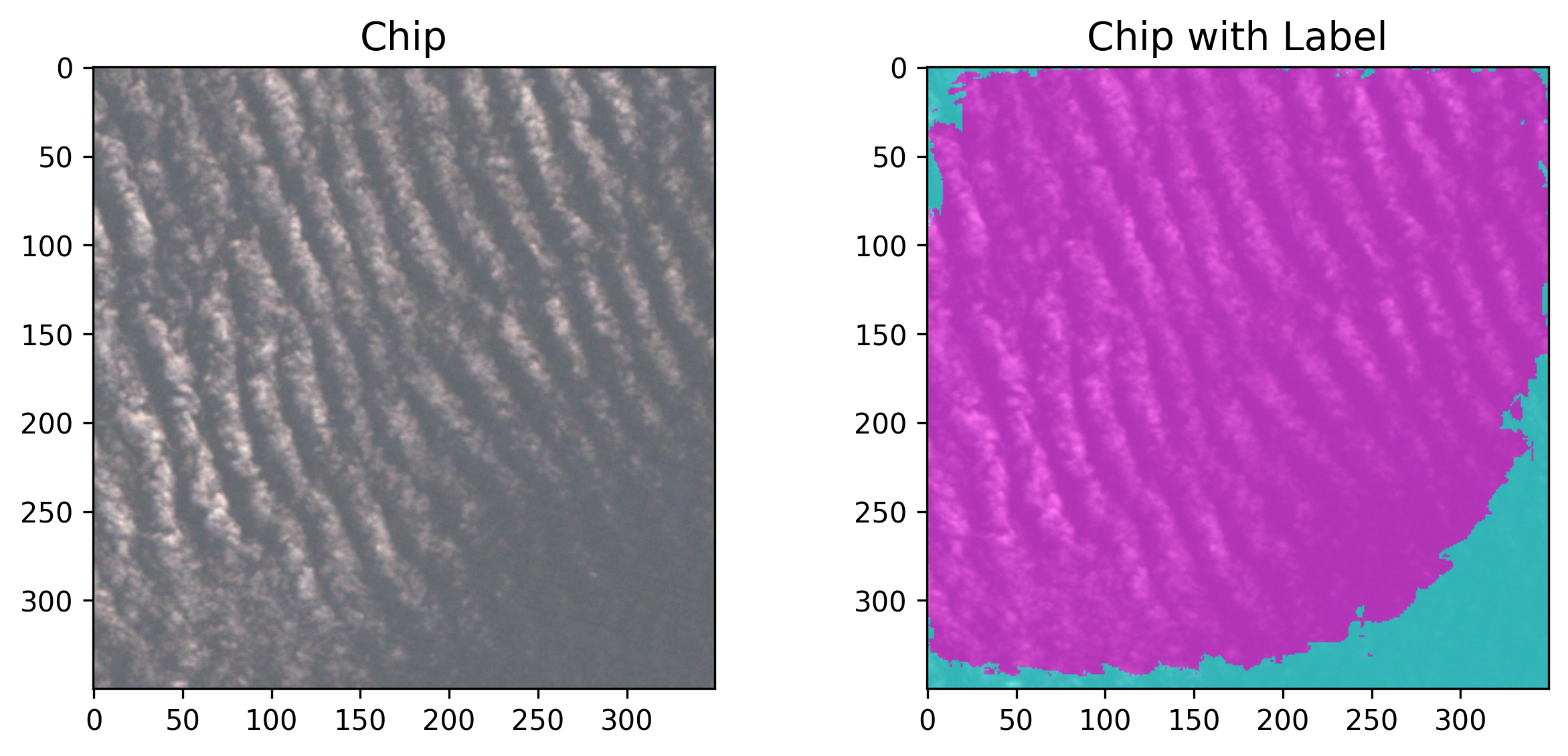}
    \caption{Annotation error, chip WU193724}
    \label{fig:miss_annotation}
\end{figure}

A quick manual review of the training images revealed that some of them contain significant annotation errors. These errors belong to two  main categories. First, some areas in a few images were marked as containing kelp forests without any kelp forest really existing. An example of this can be seen in \autoref{fig:miss_annotation}, where almost all of the image is marked as containing kelps but as we can see no kelp canopy exists and the whole image seems to be from the open sea. This is an extreme example and not many images are like this but errors in smaller scale may be more frequent.

\begin{figure}[h]
    \centering
    \captionsetup{width=.65\linewidth}
    \includegraphics[width=0.65\textwidth,trim=0 0 0 0, clip]{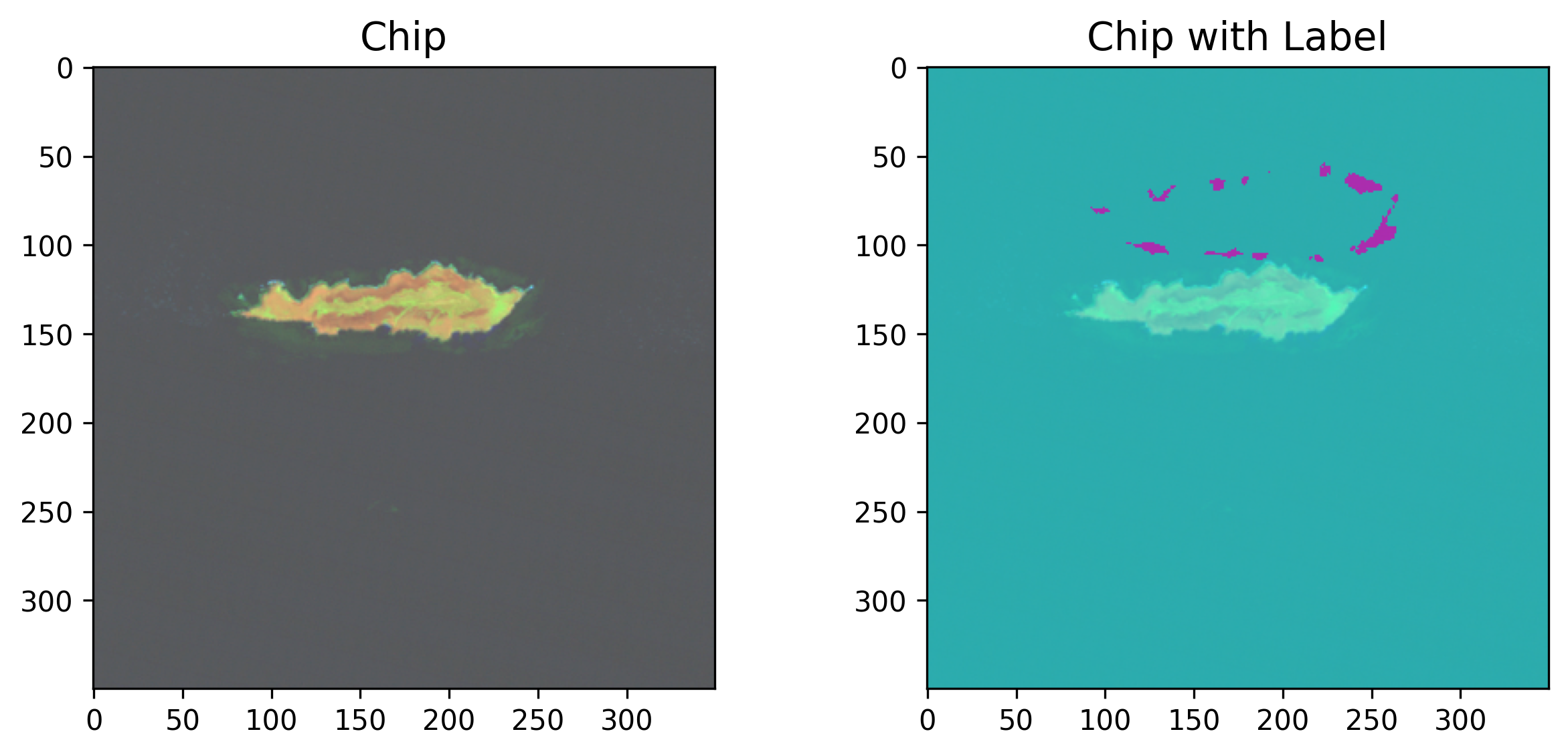}
    \caption{Shifted mask, chip QI166183}
    \label{fig:shifted_mask}
\end{figure}

The second type of miss-labelling can be seen in \autoref{fig:shifted_mask}. In cases like this the marked as area with kelp forest in regard with the true area containing kelp forests are shifted. This type of error doesn't seem possible to come from citizen scientist annotation but probably comes from the way the mask chips were cut out of a larger mask which probably differs for some cases (maybe in the edge of the large satellite product) from the way the image chips were cut out of the whole satellite product.

\subsection{Metric}
\label{sec:Metric}
To evaluate the performance in this binary semantic segmentation task, the Dice Coefficient (also known as the Sorensen-Dice Index) is used as the performance metric. The Dice Coefficient quantifies the similarity between the predicted and ground-truth binary masks. A higher Dice Coefficient indicates better segmentation accuracy.

The Dice Coefficient is calculated on a per-pixel basis, where each pixel of the predicted mask is compared to the corresponding pixel in the ground-truth mask. As the metric is calculated for the whole test dataset and is not averaged per image, it provides a more accurate measurement in kelp forest detection.

The Dice Coefficient is defined as:

\begin{equation}
\displaystyle
Dice Coefficient =   \frac{2*|A \cap B|}{|A|+|B|}  
\label{eqn:dice}
\end{equation}

where:
\begin{description}
\item[$\bullet$]  $\mid A \mid$ represents the size of set A (ground truth).
\item[$\bullet$]  $\mid B \mid$ represents the size of set B (predicted class).
\item[$\bullet$] $\mid A \cap B \mid$ represents the size of the intersection of sets A and B.
\end{description}

The above equation can also be calculated as:
\begin{equation}
Dice Coefficient = \frac{2*TP}{2*TP+FP+FN}
\label{eqn:diceF}
\end{equation}

where, TP are the True Positive, FP the False Positive and FN the False Negative pixels. This manifests the high importance of TP to be high and also for the FP and FN to be low (FN equal important to FP).

\subsection{Modelling}
\label{sec:Modelling}

\subsubsection{Data used and preprocessing}
\label{sec:prepro}
Out of 7 available input channels, only 4 of them were used. The 3 of them were used in model training while the fourth in post processing. SWIR1, NIR and Green bands were used as input to the models after preprocessing and the DEM band was used in the post processing phase by applying a land-sea mask. Training bands preprocessing includes clipping values between a minimum (6000) and a maximum (24000) value while the value of zero was set to missing data. Afterwards a division with this maximum brought all data within 0-1 range. Final step includes subtracting and dividing with imagenet's means and standard deviations (reduced) per channel. 
\begin{verbatim}
img01 = (img01 - [0.485, 0.456, 0.406])/ ([0.229, 0.224, 0.225])
\end{verbatim}

\subsubsection{Models}
\label{sec:models}
The training of segmentation models was done using the pytorch framework and specifically the segmentation\_models\_pytorch (SMP) and the transformers libraries. 
MIT and ConvNeXt models were trained in various image sizes and combined effectively. 





\subsubsection{Training parameters and augmentation}
\label{sec:params}
Training image augmentation performed using the albumentation library \citep{buslaev2020albumentations} and included vertical and horizontal flipping as well as random number of 90 degrees rotations (all with 50\% chance). A few models also included a custom augmentation for holes in both the image and the mask (chance 25\%). Images and masks are resized in various sizes (512, 640 or 768) 
using the bicubic interpolation for enlarging the images as this is preferred to bilinear and nearest-neighbor, \citep{triwijoyoa2021analysis} for segmentation tasks.

The loss function that was optimized, was the dice loss directly, which equals "1 - dice coefficient", which is the competition's metric. A custom cosine annealing scheduler with 30-35 epochs in 1 snapshot and 0-2 additional epochs for warming up, was used for all models. The U-Net segmentation models needed about 5 epochs more training than the UpperNet. The learning rate dropped from 0.00005 to 0.0000005 over these epochs while 0.000001 used at the warming up phase. In all U-Net decoders, this learning rate was always 10 times higher, meaning that in U-Net models, encoder and decoder trained with different learning rates. Finally, the AdamW optimizer was used with the default weight decay and all other parameters \citep{loshchilov2017decoupled}.

\subsubsection{Validation scheme}
\label{sec:validation}
An 80-20\% single train-validation split was used instead of a more precise k-fold split to prioritize experimentation speed. This speed-accuracy trade-off is more feasible for segmentation tasks than for classification or regression, as the large number of pixels in segmentation provides an effectively larger validation set. This split was not random but as original satellite images was not given, an attempt towards a grouped split was made utilizing custom jigsawing. This jigsawing attempt, was initially meant for augmentation purposes but as experiments didn't turn out to be as expected only used for the local validation split. For every experiment regardless the image training size the reported scoring was done in the original image size. For all experiments, scoring was reported on the original image size, while training loss was calculated on resized images, optimizing for later model ensembling across various training sizes.

\subsubsection{Final predictions}
\label{sec:predictions}
Individual model predictions on the test dataset was done with test time augmentation, using simple average of all 4 possible flip states. The final predictions was the average of the individual models predictions. These predictions are probabilities as the sigmoid function was used at the end. This probabilities turn to binary masks by applying a 0.43 threshold. Finally, the land-sea mask was applied to erase possible predicted masks on land.

\section{Results and discussion}
\label{sec:Results and discussion}

The MIT model family contain the encoders from the SegFormer model \citep{xie2021segformer} pretrained on the Imagenet \citep{deng2009imagenet}. These transformer encoders proved highly efficient for the task and outperformed many other encoders tried in local validation. Furthermore, the U-Net decoder used here \citep{ronneberger2015u} was the best as it scored better than PSPNet, FPN and MANET decoders locally, as seen in \autoref{table:cval_score_arch}. MANet models performed quite well too and ensembling U-Net with MANet predictions initially improved local results but at the final ensembling where many U-Nets were ensembled together, including MANet(s) decreased overall performance. Many other segmentation architectures were not available for the MIT encoders and therefore no comparison was possible. The overall solution improved upon ensembling predictions from the MIT U-Net models with those of the ConvNeXt encoder models \citep{liu2022convnet}. ConvNeXt models used the UpperNet decoder \citep{xiao2018unified}, increasing the variation of the decoders used and increasing the ensembling score. 

\begin{table}[ht!]
\captionsetup{format=plain, width=.5\textwidth}
\renewcommand{\arraystretch}{1.3}
\caption{Segmentation Architectures. Validation score of mit\_b1 backbone (no TTA)}
\label{table:cval_score_arch}
\centering
\begin{tabular}{l  c  } 
 \hline
 Architecture & dice coefficient \\ [1ex] 
 \hline
 U-Net & 0.7001  \\ 
 PSPNet & 0.6866  \\
 FPN & 0.6959  \\
 MANet & 0.6984  \\ 
 
 \hline
\end{tabular}
\end{table}

\begin{table*}[ht!]
\renewcommand{\arraystretch}{1.3}
\caption{Validation Scoring}
\label{table:boost_score}
\centering
\begin{tabular}{l  c  c } 
 \hline
 Model & Size & Dice Coefficient \\ [1ex] 
 \hline
 mit\_b2 & 768 & 0.708  \\
 mit\_b4 & 512  & 0.709  \\ 
 mit\_b3 & 640  & 0.709  \\ 
 mit\_b2 & 640 & 0.709  \\
  \hline
    all 4 mits & & 0.713  \\
   \hline
 convnext\_tiny & 768  & 0.706  \\
 convnext\_tiny & 512  & 0.699  \\
 convnext\_base & 512 & 0.702  \\
   \hline
    all 3 convnexts & & 0.708  \\
   \hline
    ALL = 5 * all4mits + 3 * all3convnexts & & 0.715  \\
   \hline
    ALL with altimetry post process & & 0.7208 \\
    \hline
    convelve2d on altimetry & & 0.7209 \\
    \hline
    setting mask threshold to 0.43 & & 0.7210 \\
 \hline
\end{tabular}
\end{table*}

\autoref{table:boost_score} shows that the MIT models are performing a little better than the ConvNeXt but their ensemble is even better than the combined MIT models alone. Ensembling four MIT models improved the validation score by 0.004 units and by incorporating the ConvNeXt models further improved the score by 0.002. Using the raw altimetry data at the post processing phase, to remove predicted masks on land, gave a high score boost, +0.0058. Passing the altimetry data through a convolved2d function (2*2 pixels, array of ones), smoothed the altimetry and gave a further minor improvement of +0.0001. It is unclear whether this improvement is due to minor misalignments between Landsat and altimetry data, a lack of kelp forests in tidal zones, or some other factor. Finally, further tuning the threshold for converting probabilities to mask, from 0.46 to 0.43, gave also an additional minor improvement equal to +0.0001.

\begin{table}[ht!]
\renewcommand{\arraystretch}{1.3}
\captionsetup{format=plain, width=.4\textwidth}
\caption{Validation score for various backbones - Early experiments}
\label{table:cval_score_bb}
\centering
\begin{tabular}{l  c  } 
 \hline
 Backbone & dice coefficient \\ [1ex] 
 \hline
 mit\_b1 & 0.6975  \\
 EffNetB1ns & 0.6746  \\
 RegNety064 & 0.6829  \\
 seResNeXT50\_32x4d & 0.6658  \\ 
 xception & 0.6728 \\ 
 
 \hline
\end{tabular}
\end{table}

From early experiments of \autoref{table:cval_score_bb}, it was obvious that the MIT model family was far better than the others. Therefore experiments after this initial search for the best encoder was conducted mainly by using the MIT encoders. From other encoders tested, RegNety \citep{radosavovic2020designing}, performed quite well and at first blended favourable with MIT models but didn't make it to the final ensembling as couldn't improve performance when combined with many MIT and ConvNeXt models. 

As from \autoref{sec:Annotation errors} above there are some annotation errors in data labels. Measuring performance in cleaner validation images, in 1119 out of 1128 (99.2\%), leaving 9 images out ('QI166183', 'ED338157', 'UM703003', 'OE173822', 'JD667551', 'SH612997', 'FH847016', 'ER356842', 'DU589187'), scored increased from 0.7210 to 0.7249 (0.39\% better score). This indicates that current models perform a little better than scores define. 
Using a curated dataset, following the initial crowdsourced version, is crucial for enhancing data accuracy and consistency, reducing noise, and ensuring increased model performance. Refined initial labels and addressed misclassifications, can provide a high-quality foundation for training advanced algorithms, ultimately leading to more precise monitoring outcomes.

\begin{figure*}[h]
    \centering
    \captionsetup{width=.99\linewidth}
    \includegraphics[width=0.99\textwidth,trim=0 0 0 20, clip]{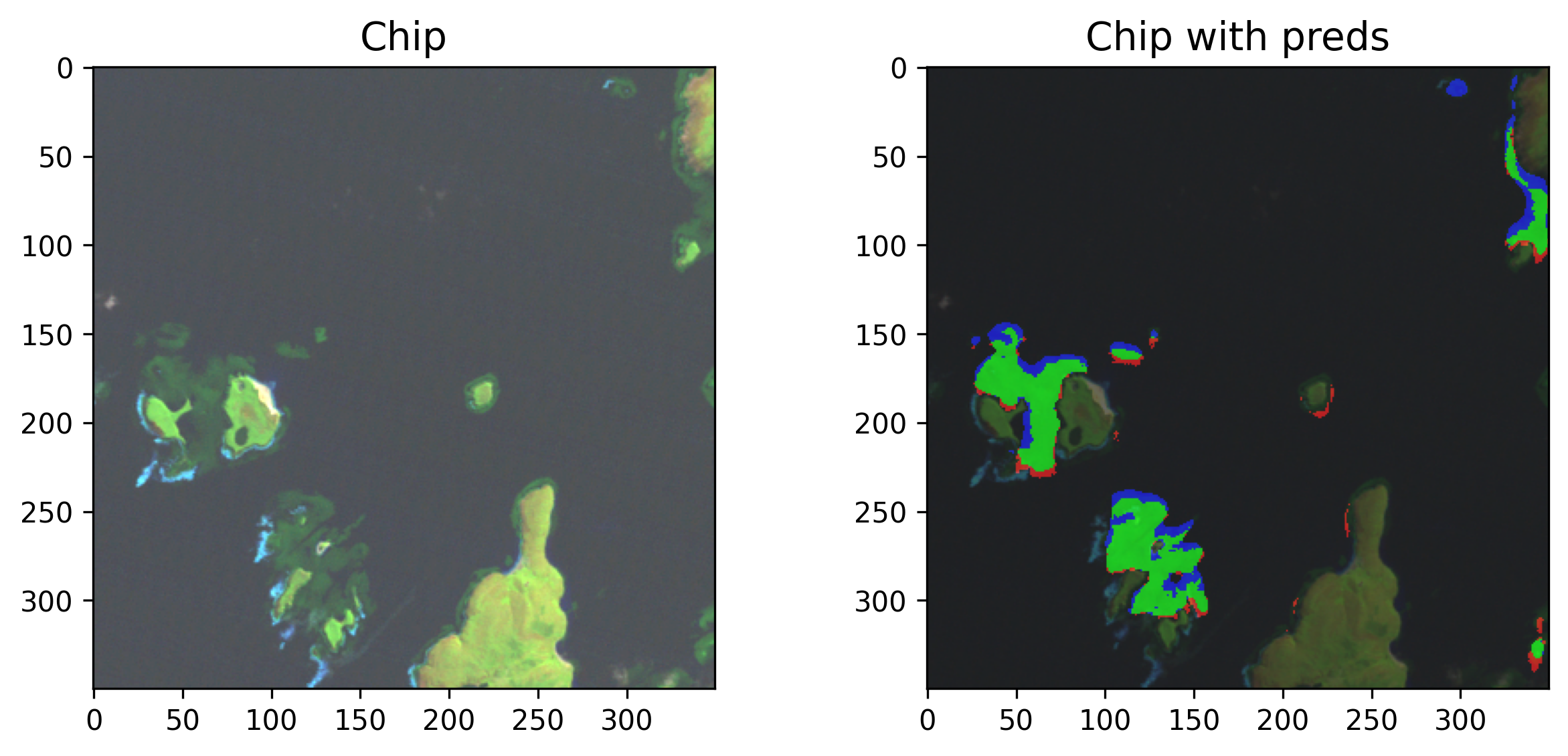}
    \caption{Image Chip on the left, image with predictions on the right. (Green=TP, Red=FP, Blue=FN)}
    \label{fig:img_preds}
\end{figure*}

As shown in \autoref{fig:img_preds}, most of the kelp forest areas are accurately predicted. However, some False Positive (FP) areas appear where the model incorrectly predicts kelp forests, and False Negative (FN) areas where existing kelp forests are not detected. These misclassified areas often occur near the edges of kelp forests, highlighting the need for enhanced model attention on boundary regions. Training with larger image sizes, as implemented in this study, helps improve predictions in these edge cases. However, further improvement using even larger image sizes is limited by the misannotations that arise from resizing the ground truth masks.

Test set is about 26\% larger than the local validation set as it contains 1426 chips while the local validation contains 1128. For the local validation set, the vast majority of pixel masks are 0s, meaning that areas with kelp forests are very small. Specifically, 855696 mask pixels are with value 1 while there are 138180000 total pixels, covering an area of about 0.62\%. Also, 436 chips contain no positive pixel at all. In another way of studying the method's performance, \autoref{table:val_pixels} shows that the majority of pixels are correctly identified (TP+TN=99.65\%) but this mainly comes from the vast empty areas. In the areas containing kelp forests about 3 out of every 4 pixels are correctly classified. 

False negatives are approximately equal to false positives, indicating that the extent of undetected kelp forests matches the areas incorrectly identified as kelp forests. Since a land-sea mask was applied, both cases are mistaken kelp forests with sea. This frequently occurs near the boundaries of predicted kelp canopies. Instances of undetected kelps or cases where a sea area has been incorrectly predicted as kelp, are less frequent and due to their small size are not affecting significant the method's performance. Some of these errors may be attributed to data misannotations rather than true inaccuracies. Finally, the resolution of the Landsat images probably have a significant role in the occurrence of both missannotations and also in model's performance at kelp boundary cases.

\begin{table}[ht!]
\renewcommand{\arraystretch}{1.3}
\caption{Validation pixels}
\label{table:val_pixels}
\centering
\begin{tabular}{l  c  c c } 
 \hline
  & Num Pixels & Total pixel percent & percent of Kelps \\ [1ex] 
 \hline
 True Negative  & 137071008 & 99.197  &  \\
 True Positive  & 625286    & 0.453   & 73.073 \\ 
 False Negative & 230410    & 0.167   & 26.927 \\ 
 False Positive & 253296    & 0.183   &   \\
 \hline
\end{tabular}
\end{table}

Dynamic environments like kelp forests require continuous monitoring to study their behavior and understand the factors influencing their distribution. Citizen science initiatives, such as the ``Floating Forests" project, make significant contributions to this effort. However, with the constant influx of satellite data, manually annotating all imagery becomes practically impossible. Utilizing artificial intelligence models trained on citizen science labels and satellite data offers an effective solution for detecting kelp forests as soon as satellite products become available, eliminating the need to wait for manual annotations. Additionally, these AI-driven predictions provide more consistent monitoring of kelp forest changes over time, as they are not reliant on varying contributions from different citizen scientists.

\autoref{table:competition_results} shows the competition results where current solution ranked in the 3rd position. Out of many experiments run, described solution here scored best in local validation as well as in public and private LB, underscoring its robustness. Scores on the public and private test sets were slightly higher than on local validation, likely due to the larger training set used for test predictions, as all available training data (train and val) was included, whereas only 80\% was used for validation. 

\begin{table}[ht!]
\renewcommand{\arraystretch}{1.3}
\caption{Competition Results}
\label{table:competition_results}
\centering
\begin{tabular}{l  c  c c} 
 \hline
 Rank & Participant & Score & public LB\\
 \hline
1 & Epoch IV	 &   0.7332 &   0.7200 \\  	
2 & xultaeculcis  &  0.7318  &  0.7197 \\  	
\textbf{3} & \textbf{ouranos}       &  \textbf{0.7296}  & \textbf{0.7247} \\  	
4 & yurithefury    & 0.7278  & - \\  	
5 & ouahab7        & 0.7264  & -  \\  	
6 & nizhib         & 0.7263  & -  \\  	
 \hline
\end{tabular}
\end{table}

In post-competition reports, the top-ranked solutions utilized various techniques to achieve their results. The first-place team trained models using calculated indices like NDVI, NDWI, and ONIR, and applied a slight Gaussian blur to targets. Additionally, they 
employed boundary loss, and used VGG-based U-Net, Swin Transformer (SwinUNetR), and ConvNeXt models. Their final solution was a weighted average of multiple models with TTA 8, involving all four flips and four consecutive 90-degree rotations. The second-place team’s approach involved averaging 10 folds using the best epoch's weights without TTA, with all training performed using U-Net models featuring an EfficientNet\_B5 encoder. They used all seven raw channels and an additional 17 computed indexes channels, along with a custom weighted sampler and quantile normalization.

Our dataset included five Landsat satellite bands, and various experiments were conducted to determine the optimal combination of bands for model input. Early results clearly showed that omitting either the NIR or SWIR bands led to a significant decline in performance. Adding the green band to the NIR and SWIR bands provided marginal improvement over using the red or green bands alone. The use of Near-Infrared (NIR) and Shortwave Infrared (SWIR) bands is crucial for accurate vegetation monitoring, especially in complex coastal ecosystems like kelp forests. These bands capture spectral signatures related to chlorophyll, moisture, and biomass, enabling effective differentiation of vegetation from other features. In our state-of-the-art AI models, incorporating NIR and SWIR bands markedly enhanced segmentation accuracy, addressing challenges such as water turbidity and shallow depths. This approach ensures more reliable, precise mapping, offering valuable insights for the conservation and management of these vital ecosystems.

This study demonstrates the effectiveness of combining Landsat imagery with crowdsourced labels and advanced deep learning models—namely, Mixed Vision Transformers and ConvNeXt segmentation models—for detecting kelp forests in remote sensing images. By harnessing the collective knowledge of citizen scientists, this approach addresses the significant challenge of generating large-scale, expert-labeled datasets in ecological research. Our findings suggest that crowdsourced labeling can serve as a feasible alternative to traditional annotation methods, especially in cases where expert input is scarce or costly. The high performance of Mixed Vision Transformers and ConvNeXt in this study also highlights their ability to effectively capture complex visual patterns within earth observation data, offering a promising pathway for future remote sensing applications. Integrating these models with crowdsourced data presents a scalable, cost-effective solution for advancing environmental monitoring and conservation initiatives.


\subsection{Insights}
\label{sec:insights}

Many things were tried along the way in order to improve model performance. As \autoref{table:boost_score_tricks} shows, using test time augmentation with 4 all possible flips was very helpful both at individual and at ensembling models level. Attempts to use TTA with rotations of 90 degrees (either alone or combined with flips) didn't improve results. 
Converting targets to floats before resizing, making target masks to include decimal values for a few pixels, further improved results. Finally, setting grads to nan upon zero\_grad call, also gave small improvement. 
The gradient update with batch accumulation, updating not on every training batch but every 3 batches, also proved beneficial with this small batch size (2-4 depending on image resize). Had a larger GPU been used, a larger batch size could had been used also and the importance of the batch accumulation may had been insignificant. 

Post-competition experiments revealed that certain strategies, initially believed to enhance the method's performance, were not as effective as expected. Increasing the U-Net decoder's learning rate 10 times, which improved performance in smaller models and image sizes, did not yield the same results for larger models. A more conservative increase (2-3 times) might be a safer option for testing. Additionally, 
the use of cubic instead of linear resizing, may have had a marginal impact on overall performance. 

\begin{table}[h]
\renewcommand{\arraystretch}{1.3}
\caption{validation score improvements}
\label{table:boost_score_tricks}
\centering
\begin{tabular}{l  c  } 
 \hline
 tips & dice coefficient \\ [1ex] 
 \hline
 TTA (4flips) & +0.0025  \\
 mask float32  & +0.0005  \\ 
 batch accumulation  & +0.0005  \\ 
 zero\_grad(set\_to\_none=True)  & +0.001  \\
 
\hline
\end{tabular}
\end{table}

\subsection{Perspectives}
\label{sec:Perspectives}
Although extensive experimentation was conducted with this dataset, there are still opportunities to further enhance the presented approach. Future experiments could incorporate strategies from other top-ranked solutions, such as including calculated indices in the training data, applying blur to training targets, training with boundary loss, and utilizing a custom weighted sampler or quantile normalization. These ideas could potentially improve the effectiveness of the method.

While the trained models can be effectively applied in real-world scenarios, ideally, they should be retrained using a curated dataset for optimal performance. This could be achieved, not necessarily by reannotating the entire dataset but by excluding images with significant annotation errors. Moreover, if the complete, uncut satellite data had been available, techniques like random crop resize could have artificially increased the training dataset, potentially enhancing model performance. Additionally, if the used for training image size of 350x350 pixels was expanded it could further improve results, as larger images provide more contextual information. Finally, as this is a relative small and from a specific area, using an even larger and more diverse dataset (including other regions, not just Falkland islands) could increase method's performance. 


The inclusion of higher-resolution satellites, such as the Copernicus Sentinels, could enhance the detection of kelp forests. However, their shorter historical record limits their effectiveness for long-term study. Exploring a combination of both Landsat and Sentinel satellites could be valuable. \citet{ed2020recent} found that using these sources together allows researchers to improve operational classification and change detection, offering deeper insights into landscape and intrinsic processes, such as deforestation and agricultural expansion.

The integration of these advanced models, paired with state-of-the-art training techniques applied to meticulously processed earth observation data, has greatly improved the accuracy and quality of kelp forest area predictions. While this approach has proven effective within the remote sensing domain, similar results may vary across other fields. However, given that these models have also shown strong performance with other types of datasets, their application is promising for a broad range of segmentation tasks. Consequently, testing these models in the experimentation phase of any segmentation project is highly recommended to assess their potential for delivering precise and reliable outcomes.

As relative research highlighted, wave dynamics and winter conditions are essential for the survival and health of kelp forests. Incorporating weather data, such as wave height, as inputs could significantly enhance our methodology’s performance in monitoring these ecosystems. By integrating various environmental factors into our model, we could gain a more nuanced understanding of how specific weather conditions impact kelp forest resilience, growth, and spatial distribution. This approach not only promises improved detection and segmentation accuracy but also could provide detailed insights into the relationship between weather variability and kelp forest health, offering valuable information for ecosystem management and conservation efforts.


\section{Conclusions}
\label{sec:Conclusions}

Kelp forests are vital ecosystems that support diverse species and play a key role in fisheries, making their conservation essential. This study demonstrates that combining Landsat imagery and crowdsourced labels with advanced machine learning models significantly enhances the accuracy of kelp forest detection in remote sensing images, providing a scalable, cost-effective solution for monitoring these habitats. Crowdsourced labeling emerges as a viable alternative to traditional annotation methods. Notably, U-Net models with MIT encoders performed exceptionally well for kelp mapping, while UpperNet architectures with ConvNeXt encoders also delivered strong results. Integrating predictions from both model types and blending outputs from models trained on varied image sizes further optimized performance. The analysis focused on three key Landsat bands—SWIR1, NIR, and Green—while altitude data was used to create a land-sea mask, reducing false positives on land. Using Level-2 data across different Landsat series eliminated the need for intercalibration while Landsat’s resolution and 40-year archive make it well-suited for long-term kelp forest studies. Although the methodology proved effective for monitoring kelp forests, accuracy could further improve with a larger, more precisely labeled dataset or the application of this approach to higher-resolution satellite imagery. This combination of advanced models and training techniques has not only enhanced kelp forest prediction accuracy but also shows promise for other segmentation applications.

\section*{Acknowledgements}
Would like to thank all contributors to the creation of both the dataset and the competition. Also thank, for the insightful commentary and suggestions, to two anonymous reviewers, which improved the quality of this manuscript. 


\section*{Declarations of conflict of interest}
The author declared that  have no conflicts of interest to this work.

\section*{Data availability}
\label{sec:Data availability}
Data are available from sources as described in \autoref{sec:Data}. Metadata as well as train labels files are available for download at competition's website, \url{https://www.drivendata.org/competitions/255/kelp-forest-segmentation/data/}



\bibliography{refs}

\end{document}